\documentclass[preprint,12pt]{elsarticle}

\usepackage{amssymb}
\usepackage{amsmath}

\usepackage{amssymb}
\usepackage{graphicx}
\usepackage{amsmath}
\usepackage{amssymb}
\usepackage{bm}
\usepackage{algorithm}
\usepackage{algpseudocode}
\usepackage{multirow} 
\usepackage{adjustbox}
\usepackage{xcolor}
\usepackage{url}
\usepackage{color,soul}
\usepackage{subcaption}
\usepackage[normalem]{ulem}

\journal{Nuclear Physics B}

\begin{document}

\begin{frontmatter}



\title{Hybrid Robot Learning for Automatic Robot Motion Planning in Manufacturing} 


\author[1]{Siddharth Singh\corref{contrib}} 
\ead{sks4zk@virginia.edu}
\author[2]{Tian Yu\corref{contrib}} 
\ead{Tian.Yu@geaerospace.com}
\author[1]{Qing Chang\corref{corauthor}} 
\ead{qc9nq@virginia.edu}
\author[2]{John Karigiannis}
\ead{John.Karigiannis@geaerospace.com}
\author[2]{Shaopeng Liu}
\ead{sliu@geaerospace.com}

\cortext[corauthor]{Corresponding author}
\cortext[contrib]{Authors contributed equally}

\affiliation[1]{organization={Mechanical \& Aerospace Engineering, University of Virginia},
addressline={122 Engineer's Way}, 
city={Charlottesville},
postcode={22903}, 
state={VA},
country={U.S.A}}

\affiliation[2]{organization={General Electric Aerospace},
addressline={122 Engineer's Way}, 
city={Nisskayuna},
postcode={22903}, 
state={NY},
country={U.S.A}}

\begin{abstract}
Industrial robots are widely used in diverse manufacturing environments. Nonetheless, how to enable robots to automatically plan trajectories for changing tasks presents a considerable challenge. Further complexities arise when robots operate within work cells alongside machines, humans, or other robots. This paper introduces a multi-level hybrid robot motion planning method combining a task space Reinforcement Learning-based Learning from Demonstration (RL-LfD) agent and a joint-space based Deep Reinforcement Learning (DRL) based agent. A higher level agent learns to switch between the two agents to enable feasible and smooth motion. The feasibility is computed by incorporating reachability, joint limits, manipulability, and collision risks of the robot in the given environment. Therefore, the derived hybrid motion planning policy generates a feasible trajectory that adheres to task constraints. The effectiveness of the method is validated through simulated robotic scenarios and in a real-world setup. \textit{Project Website:\url{https://sites.google.com/virginia.edu/isl-hllfdra/home}}

\end{abstract}




\end{frontmatter}



\section{Introduction}
In today's manufacturing settings, there is a clear shift towards agile, intelligent production with an enhanced role for robots \cite{lo22}. Yet, the industry predominantly leans on pre-programmed robots, necessitating reprogramming even for minimal task adjustments. The associated time and costs in reprogramming these robotic systems present notable challenges \cite{lrfchc21}, \cite{zlcwg20}.

To address challenges in robot motion planning, researchers have developed two main approaches: joint space and task space planning. Joint space planning avoids singularities—including joint limits and obstacles—allowing precise control over individual joints for detailed motion planning \cite{yo20}, \cite{yjwl21}. However, it is usually time-consuming and lacks task-level understanding, leading to potential inaccuracies in the end-effector's motion. Task space planning, on the other hand, focuses on task-centric operations. It is fast and adaptable to different robots but relies on inverse kinematics for joint angle calculations, posing challenges in avoiding singularities, collisions, and joint limits \cite{km22}, \cite{amr22}. Hence, a method combining the merits of both approaches while mitigating their limitations is essential.

In recent years, with the development and advancement of Artifical Intelligence (AI), Deep Reinforcement Learning (DRL) has become crucial for robot motion planning \cite{hmsc23}. For instance, the authors propose a Soft Actor Critic (SAC) based method \cite{ko22} for self-homing in industrital robotic cell. This method assumes a pre-sensed unknown environment, allowing for policy transfer without extra training. It employs a multi-agent training setting, enhancing state space exploration, with agents sharing experiences and deploying policies collectively. While DRL methods like Deep Q-network (DQN) \cite{lzhggf22}, \cite{zjwzt22}, Deep Deterministic Policy Gradient (DDPG) \cite{km22}, \cite{vcsh21}, and Proximal Policy Optimization (PPO) \cite{ftcg22} are effective for tasks like pick-and-place and peg-insertion, these methods also face challenges. These challenges include difficulty learning implicit task constraints, requiring extensive data and time, and lacking consistent stability and accuracy. Modifications to work cells or tasks also demand DRL agent retraining or finetuning \cite{zo22}, highlighting the need for ongoing refinement in DRL techniques.
Compared with modeling a task and planning the motion of a robot in the DRL-based robot motion planning, human operators are often more intuitive in performing the task. Learning from Demonstrations (LfD) offers a viable strategy for robots to execute similar tasks in Human-Robot Collaboration environments due to its intuitive nature. However, LfD methods currently face challenges with scalability and adaptability \cite{mgcn22}, \cite{yc22} The authors have introduced a scalable LfD technique, allowing robots to devise adaptive motion plans from a single demonstration \cite{rpcb20}. Although this approach efficiently maps kinematic features to new tasks, its efficacy is limited in complex environments with varied obstacles. Robots often require search algorithms to navigate, making original demonstrations less applicable. Furthermore, task space LfD algorithms may encounter joint-space issues, including self-collisions, reachability, and manipulability limitations \cite{bwb20}.

To address these difficulties, this paper introduces a multi-level hybrid robot motion planning method, integrating the task space LfD and joint space DRL-based approaches, aiming to capitalize on their respective strengths while mitigating their limitations. Our multi-level approach incorporates three distinct learning based agents. First a Hierarchical Reinforcement Learning-based LfD (RL-LfD) method is trained to utilized the skill library to generate a trajectory in task space. Then a DRL based agent is trained to overcome infeasible joint trajectories. Lastly, an RL based switching agent is trained, which learns to identify an optimal switching strategy between the skill-based agent and the DRL agent to maximize the feasibility and ensure smooth transition between the trajectories generated by the two agents. The main contributions of this work are:
\begin{enumerate} 
    \item \textbf{Feasible Learning based Motion Planning}: We leverage both human-demonstrations and DRL to generate motion plans while taking the feasibility factors such as reachability, manipulability, collision-check, into account.  
    \item \textbf{Multi-level Motion Planning}: A multi-level motion planning approach is proposed which includes combining the learning based agents and also learning appropriate switching behavior to generate smooth and safe joint trajectories.
    \item \textbf{Implicit adherence to task constraints}: We further augment LfD agent to devise motion plans for a long-horizon tasks while maintaining joint feasibility ensuring the adherence to task constraints.
\end{enumerate}


The remainder of this paper is organized as follows: Section \ref{sec:math_back} entails the mathematical background. In Section \ref{sec:prob_form}, the motion planning problem is formulated, and the hybrid motion planning method is introduced. Section \ref{sec:feas} introduces the joint space feasibility analysis. The task space HRL-based LfD method and the joint space DRL-based method are discussed in Section \ref{fig:HRL_Lfd} and \ref{sec:drl_pla}. Numerical case studies are provided in Section \ref{sec:exp}. Section \ref{sec:conc} summarizes the conclusion and future work.

\section{Mathematical Background} \label{sec:math_back}
\subsection{Mathematical Notations}
We define the $n-$dimensional Euclidean space as $\mathbb{R}^n$. Bold letters are used to define vectors, e.g. $\mathbf{v}$. Matrices are presented by bold capital letters such as $\mathbf{A}$. A set is represented by curly capital letters, e.g. $\mathcal{A}$. The Special Orthogonal Group is represented as $SO(3)$ which is the space of rigid rotations in 3-D space, and the Special Euclidean Group is represented as $SE(3)$ which is the group of rigid transformations in 3-D space.

\subsection{Dual Quaternion}
In this work we define the demonstrations in $SE(3)$ space. To describe the configurations, we utilize a dual quaternion encoding both the rotational and the translational information. In this paper a dual quaternion is defined as 

\begin{equation}\label{eq:dual_quatdef}
    \mathbf{q} = \mathbf{q}_r + \frac{1}{2}\eta(\mathbf{q}_t \otimes \mathbf{q}_r)
\end{equation}

where $\eta \neq 0$, but $\eta^2=0$ and $\otimes$ represents the quaternion multiplication. In eq. 2, $\mathbf{q}_t$ is the quaternion representing the pure translation of the rigid body, expressed as 

\begin{equation}
    \mathbf{q}_t = (0, \hat{\mathbf{t}})
\end{equation}

where $ \hat{\mathbf{t}} = x \hat{\mathbf{i}} + y  \hat{\mathbf{j}} +  z \hat{\mathbf{k}}$ representing the translation in $SE(3)$. Similarly, in eq.\:\ref{eq:dual_quatdef}, $\mathbf{q}_r$ represents the rotational orientation of the rigid body which is defined as:

\begin{equation}
    \mathbf{q}_r = cos(\frac{\phi}{2}) + \hat{\mathbf{v}}sin(\frac{\phi}{2})
\end{equation}

where $\hat{\mathbf{v}} = v_x \hat{\mathbf{i}} + v_y  \hat{\mathbf{j}} +  v_z \hat{\mathbf{k}}$, is the unit vector in $SE(3)$ along the axis of rotation and $\phi$ is the angle of rotation. 

\subsection{Markov Decision Process Formulation}
A Markov Decision Process (MDP) is a task which satisfies the Markov property; represented by 4 tuple, generally represented as $(\mathcal{S},\mathcal{A},P,R)$. Here $\mathcal{S}$ represents the set of states called the state space. $\mathcal{A}$ is the set of actions, $P(s' | s,a) : \mathcal{S} \times A \times \mathcal{S} \rightarrow [0,1]\;  \forall \; s,s' \in \mathcal{S},a \in \mathcal{A}$,  is a probability function which determines the probability transitions to a new state, $s'$ given a state action pair $(s,a)$. And $R: \mathcal{S} \times \mathcal{A} \rightarrow \mathbb{R}$ is the function that maps the state-action pair to a scalar value which is referred as the reward. A policy $\pi(a|s): \mathcal{S} \times \mathcal{A} \rightarrow [0,1] $ is a function that maps an action $a\in \mathcal{A}$ given a state $s\in \mathcal{S}$ to a probability distribution. If the policy maps the state to a particular action, then it is defined as a deterministic policy and if the mapping leads to a probability distribution, then it is defined as a stochastic policy. In our case, we assume that the policy maps the given state to a distribution, hence our policies are stochastic. The goal of the agent is to learn the policy that maximizes the expected cumulative reward. If the discounted cumulative reward is defined by $G = r_t + \gamma r_{t+1} + \gamma^2 r_{t+2} + \gamma^3 r_{t+3} + \cdots $, where $\gamma$ is the discounting factor, we can define a Q-function as $Q^\pi (s,a)=\mathbb{E}[G | s,a]$. Hence, the policy associated to the Q-function is given by 
\begin{equation}
    \pi(a|s) = \arg \max_{a} Q^{\pi}(s,a)
\end{equation}

Accordingly, the optimal policy is defined as:
\begin{equation}
    \pi^* :=  \arg \max_{\pi}\left( Q^\pi (s,a) = \mathbb{E}[G|s,a]\right)
\end{equation}

\section{Hybrid Motion Planning Framework}\label{sec:prob_form}
In this paper, we delve into a scenario where a robotic manipulator is tasked with performing distinct assignments, such as material handling, painting, assembly, or inspection \cite{bwb20}. Depending on the particular task at hand, various constraints are typically imposed, encompassing factors like initial and target positions, as well as the requirement of specific orientations for the end effector. In manufacturing environments, work cells often exhibit intricate layouts comprising machines, robots, and various components. It is assumed that the position and dimension of the workspace and each component in the workspace are observable. Also, the critical configurations of the new task are given. This research aims to create an effective automated motion planning solution for a robot manipulator, thereby mitigating the need for costly reprogramming when executing diverse tasks within a complex industrial environment. 

The authors have previously developed two approaches with the same goal. The task space-based robot LfD approach \cite{yc22} is effective at quickly and accurately learning from a single demonstration but struggles in complex robot environments with various obstacles, requiring search algorithms for navigation. Additionally, it operates in task-space motion planning, potentially encountering issues in joint-space. In contrast, the DRL-based motion planning approach \cite{bwb20} guarantees joint-space solutions but demands substantial data for agent training, involves time-intensive computations, and lacks inherent task-specific constraint adherence.

\subsection{Problem Formulation}
For a 3-D workspace, denoted as $\boldsymbol{WS}$, joint-space feasibility is estimated, resulting in the formation of a discrete feasibility map, $\boldsymbol{CM} \subset \boldsymbol{WS}$, where each specific pose in $\boldsymbol{CM}$ denotes an end effector configuration, including both position and orientation. $\boldsymbol{CM}$ is divided into two primary regions, the feasible region, $\boldsymbol{FR}$, and infeasible region, $\neg \boldsymbol{FR}$, which correspond to the joint-space analysis of each end effector configuration. For $\neg \boldsymbol{FR}\subset \boldsymbol{CM}$, three potential conditions may arise: the manipulator's end effector might be unable to reach a pose, a self-conflict or collision with workspace obstacles may occur, or the manipulator's manipulability \cite{y85} (a measure of how close a manipulator is to singularity) might diminish. Therefore, $\boldsymbol{FR}\subset \boldsymbol{CM}$ represents the desirable region for motion planning.

Subsequently, a feasibility map $\boldsymbol{CM}$ is utilized as one of the deciding criterion for the trajectory generated by the task-space robot RL-LfD method (will be introduced in Section \ref{sec:hrl_lfd}), referred to as $\text{traj}_{\text{LfD}}$.
Subsequently, the feasibility map $\boldsymbol{CM}$ and the feasibility of the generated motion plan is utilized to assess the performance of the executed trajectory. It is important to note that $\text{traj}_{\text{LfD}}$ may encompass segments which are infeasible or pass through the infeasible region. The feasible segments can be represented as $\boldsymbol{FJ} \subseteq \{ \text{traj}_{\text{LfD}} \cap \boldsymbol{FR} \}$, while the infeasible segments can be represented as $\neg \boldsymbol{FJ} \not\subseteq \{ \text{traj}_{\text{LfD}} \cap \boldsymbol{FR} \}$. For the segments falling within $\boldsymbol{FJ}$, inverse kinematics will be calculated to obtain the joint angles to control the robot. For segments within $\neg \boldsymbol{FJ}$, a DRL approach, which will be introduced in Section \ref{sec:drl_pla}, is utilized to guide these segments into the $\boldsymbol{FJ}$ category, resulting in a feasible trajectory $\text{traj}_{\text{DRL}}$. 

\subsection{Proposed Framework}
\begin{figure}
    \centering
    \includegraphics[width=1\linewidth]{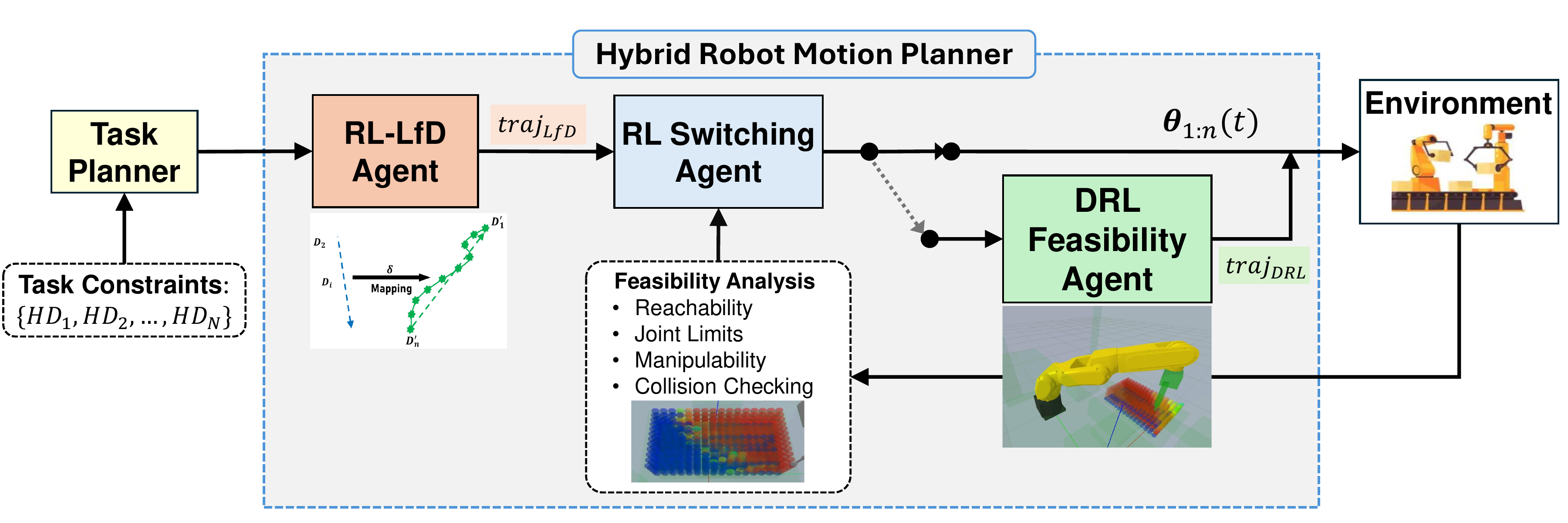}
    \caption{Framework of the proposed multi-level hybrid motion planning technique.}
    \label{fig:prob_form_sid}
\end{figure}

In order to address the significant challenge of enabling efficient robot motion planning within complex manufacturing environments and to overcome the challenges posed by both methods (i.e., task-space-based LfD and joint-space-based DRL methods), a novel hybrid approach is introduced. This method aims to systematically integrate the two approaches, utilizing the LfD method for its scalability and task space understanding, and incorporating the DRL-based approach to ensure joint space feasibility. The framework of this approach is shown in Fig.\ref{fig:prob_form_sid}. Given the task constraints, first a skill-based LfD agent is utilized to generate motion plans, $\text{traj}_{LfD}$. For the portion of the trajectory where the trajectory generated by the motion plan is infeasible we leverage a DRL agent which emphasizes on maintaining feasibility, $\text{traj}_{DRL}$. Lastly, to identify an efficient way to switch between the two agents and generate a seamless motion we train a RL based switching agent.

As such, the RL-LfD and DRL motion planning method (see Algorithm  \ref{alg:hrl_lfd} in Section \ref{sec:hrl_lfd} and Algorithm  \ref{alg:drl} in Section \ref{sec:drl_pla}), can be systematically integrated into a hybrid motion planning approach. The offline training and online execution of the hybrid motion planning method can be shown as Algorithm \ref{alg:complete_method}. It is worth noting that, in offline training, a set of tasks are generated randomly to train the RL-LfD motion planning policy, and the data of infeasible segment $\neg \boldsymbol{FJ}$ for each $\text{traj}_{\text{LfD}}$ can be used to train the DRL policy. Since the task space RL-LfD method is a scalable and adaptive method for different tasks, when task changes, the hybrid method does not need to be retrained. In addition, the set of $\neg \boldsymbol{FJ}$ represents the region in the workspace where the infeasibility usually happens. With such knowledge obtained from RL-LfD, the DRL method does not need to learn from scratch by searching the whole workspace. This method significantly shrinks the search space and boosts training efficiency. While the two agents are independently feasible, to ensure that the final trajectory is also maintains feasibility constraints and there is a seamless switch between the two agents, we train another RL agent which observes the complete trajectory execution and learns to optimize the switching action. It is referred to as Switching agent (see Section \ref{sec:switch}). This ensures smooth transitions between the motion policies generated by the RL-LfD agent and the DRL agent. In online execution, the resulting trajectory, $\text{traj}_{\text{final}}$, derived from this hybrid approach represents an optimized path that not only satisfies the task space constraints but also is feasible in the joint space. 

\begin{algorithm}
\caption{Hybrid Method: Offline Training and Online Execution}
\label{alg:complete_method}
\begin{algorithmic}[1]
\Procedure{Offline Training of the Hybrid Method}{}
\State \textbf{Input:} Demonstration Library (\({\boldsymbol{LB}}\)), \(\boldsymbol{WS}\), Robot
\State Calculate the feasibility map (\(\boldsymbol{CM}\))
\State Initialize a set of tasks (\(\boldsymbol{TK}\)) with random starting and end positions
\For{each \(\boldsymbol{TK}\)}
    \State Call Procedure 1 in Algorithm  \ref{alg:hrl_lfd} and save each \( \neg \boldsymbol{FJ} \)
\EndFor
\State Import all \( \neg \boldsymbol{FJ} \) to Procedure 1 in Algorithm  \ref{alg:drl}

\State \textbf{Output:} Trained DQN of the RL-LfD method and DRL Agent
\EndProcedure

\Procedure{Train Switching Agent}{}
\State \textbf{Input:} New Task (\(\boldsymbol{TK}\)), Demonstration Library (\({\boldsymbol{LB}}\)), \(\boldsymbol{WS}\), Robot
\State Initialize Switching Policy \(\pi_s\)
\State Call Algorithm  \ref{alg:hrl_lfd} to calculate the task space trajectory \( \text{traj}_{LfD} \)
\For{the infeasible segment \( \neg \boldsymbol{FJ} \)}
    \State Call Algorithm  \ref{alg:drl} to calculate the joint space trajectory \( \text{traj}_{DRL} \)
\EndFor
\For{the feasible segment}
    \State Calculate inverse kinematics
\EndFor
\State \( \text{traj}_{comb} \leftarrow \) Concatenate \( \text{traj}_{DRL} \) and the joint space \( \boldsymbol{FJ} \)
\State Call switching policy, \(\pi_s\)(\(traj_{comb}\)) and compute total reward \(r_s\)
\State Update switching agent policy parameters \(\pi^*_s \leftarrow \pi_s\)
\State \textbf{Output:} \( \text{traj}_{final} \)
\EndProcedure

\Procedure{Online Execution of the Hybrid Method}{}
\State \textbf{Input:} New Task (\(\boldsymbol{TK}\)), Demonstration Library (\({\boldsymbol{LB}}\)), \(\boldsymbol{WS}\), Robot

\State Call Algorithm  \ref{alg:hrl_lfd} to calculate the task space trajectory \( \text{traj}_{LfD} \)
\For{the infeasible segment \( \neg \boldsymbol{FJ} \)}
    \State Call Algorithm  \ref{alg:drl} to calculate the joint space trajectory \( \text{traj}_{DRL} \)
\EndFor
\For{the feasible segment}
    \State Calculate inverse kinematics
\EndFor
\State \( \text{traj}_{comb} \leftarrow \) Concatenate \( \text{traj}_{DRL} \) and the joint space \( \boldsymbol{FJ} \)
\State \textbf{Input:} \( \text{traj}_{comb}\) to \(\pi^*_s\)
\State Compute optimal switching sequence 
\State \textbf{Output:} \( \text{traj}_{final} \)
\EndProcedure
\end{algorithmic}
\end{algorithm}

\section{Joint Space Feasibility Analysis}\label{sec:feas}
This section presents a thorough feasibility study that combines the evaluation of reachability, joint limits, manipulability, and collision checking for the robot, resulting in the creation of a feasibility map. This map will serve the purpose of streamlining the process of hybrid motion planning. The map is leveraged for learning the switching policy which will systematically determine whether to employ an RL-LfD or a DRL-based method to automatically generate a manipulator motion plan.  

\subsection{Integrated Feasibility Map}
As such, the RL-LfD and DRL motion planning method (see Algorithm  \ref{alg:hrl_lfd} in Section \ref{alg:hrl_lfd} and Algorithm  \ref{alg:drl} in Section \ref{sec:drl_pla}), can be systematically integrated into a hybrid motion planning approach using the feasibility map and the Switching policy. The offline training and online execution of the hybrid motion planning method can be shown as Algorithm \ref{alg:complete_method}. It is worth noting that, in offline training, a set of tasks are generated randomly to train the RL-LfD motion planning policy, and the data of infeasible segment $\neg \boldsymbol{FJ}$ for each $\text{traj}_{\text{LfD}}$ can be used to train the DRL policy. Since the task space RL-LfD method is a scalable and adaptive method for different tasks, when task changes, the hybrid method does not need to be retrained. In addition, the set of $\neg \boldsymbol{FJ}$ represents the region in the workspace where the infeasibility usually happens. With such knowledge obtained from RL-LfD, the DRL method does not need to learn from scratch by searching the whole workspace. This method significantly shrinks the search space and boosts training efficiency. In online execution, the resulting trajectory, $\text{traj}_{\text{final}}$, derived from this hybrid approach represents an optimized path that not only satisfies the task space constraints but also is feasible in the joint space.

While existing studies have discussed reachability \cite{mg18}, manipulability \cite{tmt22}, and collision checking \cite{zz22}, most of them focus on individual problems. This paper develops a holistic study to integrate all aspects and provide a feasibility map. In this paper, the configuration that integrates position and orientation of the end-effector in $SE(3)$ is represented as a dual-quaternion $D \in \mathbb{R}^8$, which is an 8-dimensional real algebra isomorphic to the tensor product of the quaternions and the dual numbers \cite{s05}. Given a configuration $D$ and the joint limits $B \in \mathbb{R}^n$, where $n$ is the degrees of freedom (DoF), the reachability of the manipulator can be determined by checking the existence of solutions to the inverse kinematics $\text{IK}: D \to \theta$ such that $\theta \in B$, where $\theta \in \mathbb{R}^n$ is a vector of joint angles. It can be defined as:
\begin{equation}
RH(D,B) = \begin{cases} 
1, & \text{if IK solutions exist} \\
0, & \text{otherwise} 
\end{cases}
\end{equation}
However, if only reachability constraint is considered some poses can be reached but can lose one DOF in translation or rotation. To tackle this problem, manipulability is considered as a second constraint. The conventional measurement of manipulability \cite{y85} is given as :

\begin{equation}
\text{man}(\boldsymbol{\theta}) = \sqrt{\det[J(\boldsymbol{\theta}) J^T(\boldsymbol{\theta})]}
\end{equation}

where $J(\theta) \in \mathbb{R}^{m \times n}$, $m$ is the DOF of the end-effector and $n$ is the DOF of the robot arm, $J(\theta)$ is the Jacobian matrix and $J^T(\theta)$ is the transpose of $J(\theta)$. This is a metric that quantifies the distance between a manipulator state $\theta$ and a singularity state $\theta^*$. Such a measurement can provide information about the overall movement ability of the end-effector. By comparing the manipulability $\text{man}(\theta)$ with the manipulability of the initial pose of the robot $\text{man}(\theta_0)$, the normalized manipulability can be defined as:

\begin{equation}
    man' = \frac{man(\boldsymbol{\theta})}{man(\boldsymbol{\theta}_0)}
\end{equation}


Furthermore, it is essential to conduct collision checks, especially considering a manipulator's potential interactions with workspace obstacles or its own components. In this paper, the collision detector in PyBullet \cite{cb16} is utilized for these collision checks. In PyBullet, the Continuous Collision Detection (CCD) library first simplifies complex objects by decomposing them into simpler convex pieces, and then identifies overlaps of pieces, which furnishes detailed insights such as contact points, contact joint or link indices, and penetration depths. For a configuration $\theta$, the collision index is denoted as $\text{COL}(\theta)$. Using CCD, if $\text{COL}(\theta) = 1$ then there is a collision identified; otherwise, $\text{COL}(\theta) = 0$. Therefore, the feasibility measurement of a configuration is found by  integrates reachability, manipulability, and collision checking.

\begin{figure}
    \centering
    \includegraphics[width=0.75\linewidth]{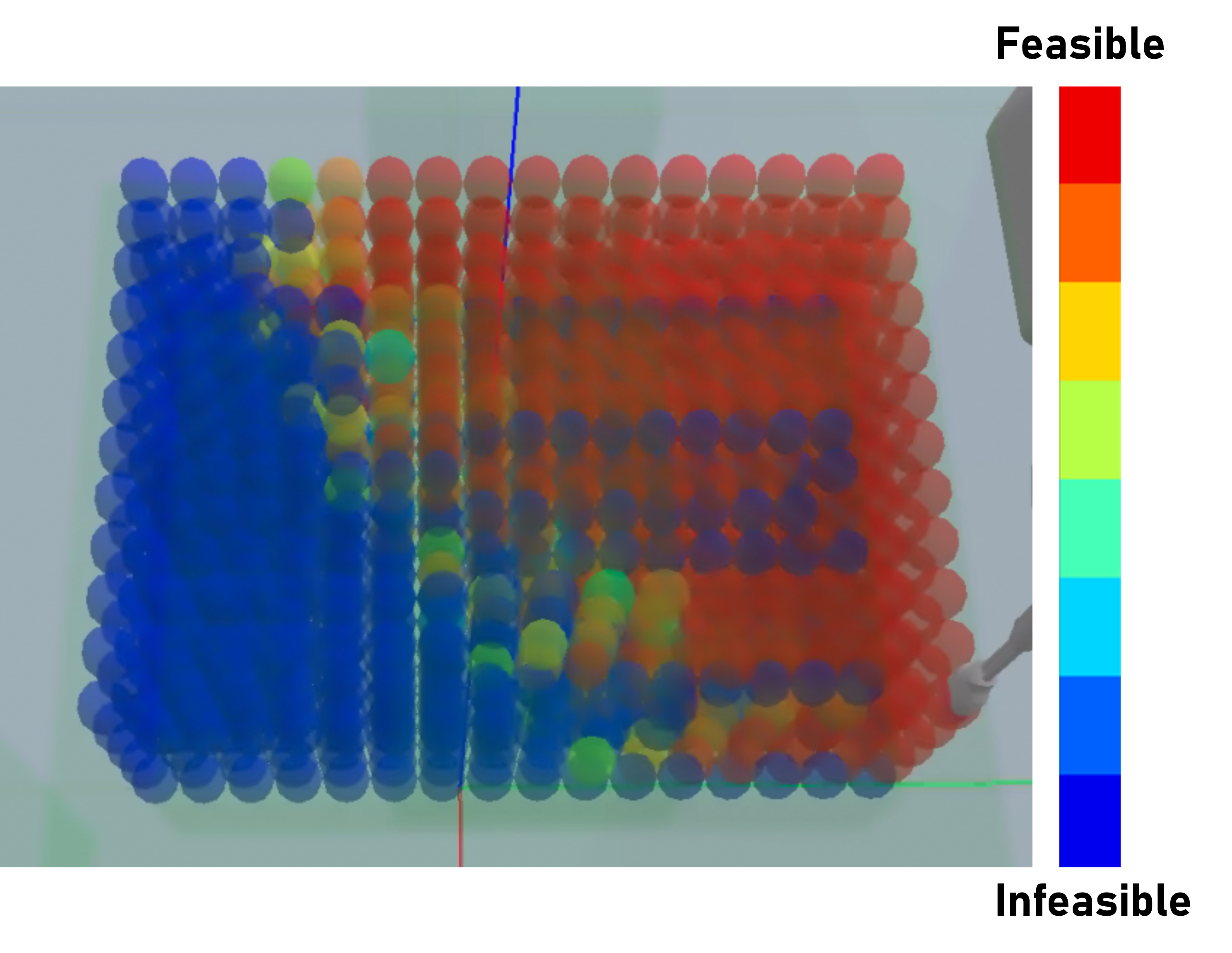}
    \caption{Schematic picture of the feasibility map in a robotic workspace.}
    \label{fig:Feas_map}
\end{figure}

To compute the feasibility of the workspace we use a tessellation  approach. Within the work cell environment $W$, it is straightforward to define a 3-D workspace, $\boldsymbol{WS}$, according to task specifications, such as the precise painting of a component at a designated location. $\boldsymbol{WS}$ can then be discretized into small voxels, each with its center signifying the $x,y,z$ position of the end effector. Within each voxel position, the orientation of the end effector in terms of $\alpha,\beta,\gamma$ can be further discretized within a task-specific range of $[-\theta,\theta] \subseteq [-\pi,\pi]$. Consequently, the discretized $WS$ forms a tensor of rank 2, where each array corresponds to a unique end effector configuration. Using Eqn. (4), it becomes possible to assess the feasibility of each end effector configuration within $\boldsymbol{WS}$. This discretized feasibility assessment can be analogized to a “map”, referred to as $\boldsymbol{CM}$, which divides the workspace $\boldsymbol{WS}$ into feasible and infeasible regions based on the criterion $\text{fea}(D)$. The map $\boldsymbol{CM}$ effectively encapsulates the end effector configuration along with its associated joint space feasibility status. Fig. 2 is a 3D illustration of a feasibility map where both feasibility and unfeasibility regions are color-coded. It is important to acknowledge that the “actual map” encompasses six dimensions, which include both position and orientation, making it impossible to visualize directly. Therefore, each voxel in Fig. \ref{fig:Feas_map} includes a specific position but includes a range of orientation that may or may not be in the feasible region, which is represented by a transition color between red (representing feasible) and blue (infeasible).

\subsection{Hybrid Motion Planning with RL based Switching Agent}\label{sec:switch}
For a task $t_k$, using the RL-LfD method, which will be introduced in the next section, a task space trajectory, $\text{traj}_{\text{LfD}} = \{D_1, D_2, \ldots, D_n\}$, can be generated. Starting from $D_1$, by checking the feasibility of each configuration, the infeasible trajectory segment, e.g., $\neg \boldsymbol{FJ}_{ij} = \{D_i, D_{i+1}, \ldots, D_j\}$, $i,j \in \{2, \ldots, n-1\}$ can be determined. The DRL method is used to compensate for each $\neg \boldsymbol{FJ}_{ij}$. 



For many tasks, such as achieving a proper grip for stacking, it is more crucial to mimic the latter part of the $\text{traj}_{\text{LfD}}$ generated through LfD, rather than the initial stages. This assumption ensures the utilization of DRL motion planning for handling the infeasible segment, $\neg FJ_{ij}$, since DRL-based methods excel at ensuring joint-space feasibility but may not adequately address task-specific constraints. In an effort to minimize $\neg \boldsymbol{FJ}_{ij}$, we anticipate that the transition from $\text{traj}_{\text{DRL}}$ to $\text{traj}_{\text{LfD}}$ should be done maximize the feasibility. To ensure that switching occurs seamlessly, at the correct time and to ensure that $\text{traj}_{comb} := \text{traj}_{\text{LfD}} ~\cup ~\text{traj}_{\text{DRL}}$ is feasible, we introduce a RL agent, referred to as the Switching Agent. To train the policy for switching action, we execute the $\text{traj}_{comb}$ and observe the award accumulated which encapsulates the feasibility of the trajectory based on the terms defined in the previous section. We define the reward as:
\begin{equation}
    r_s = \sum_{t=0}^{n_p} (man'(\theta_t) - COL(\theta_t))
\end{equation}
Here, $n_p$ are the number of points in the $\text{Traj}_{comb}$. The observation are the joint trajectories and the action space is a disrcetized one-hot encoded variable which encapsulates the choosing between the two agents. For the purpose of training the switching the agent, the LfD agent and the DRL-Feasibility agent 


It is important to note that when learning from different demonstrations for the same task, $t_k$, various trajectory outcomes, $\text{traj}_{\text{LfD}}$, can be achieved. To evaluate each $\text{traj}_{\text{LfD}}$, a criterion needs to be established. Notably, the policy generated through the proposed DRL method is not deterministic, leading to varying $\text{traj}_{\text{DRL}}$ lengths for identical tasks. On the other hand, each $\text{traj}_{\text{LfD}}$ acquired through learning from a specific demonstration is deterministic in nature. The core concept behind the suggested hybrid approach is to maximize the utilization of LfD motion planning due to its computational efficiency and the generalizability inherent in the proposed LfD method. Consequently, minimizing $\neg \boldsymbol{FJ}_{ij}$ serves as the prime criterion when selecting $\text{traj}_{\text{LfD}}$, facilitating a highly efficient fusion of $\text{traj}_{\text{LfD}}$ motion planning and $\text{traj}_{\text{DRL}}$ motion planning in our hybrid approach. It is important to note that the $\text{traj}_{comb}$ is used for training the switching agent offline.

\section{Task-Space RL based LfD}\label{sec:hrl_lfd}

In manufacturing, manipulator motion planning can be complex, but human operators can intuitively demonstrate tasks. The authors have created a task-space LfD method for robot manipulators \cite{yc22}, enabling them to learn specific tasks like object grasping or relocation based on a single primitive skill. In this paper, we expand our earlier LfD method.

First, a local motion planner is developed by using kinematic task-space planning that follows the implicit geometric constraints throughout a one-time human demonstration of a primitive skill. Then, built upon the local motion planner, a global planner is established, which can enable robots to automatically generate their motion plan for various tasks by intelligently combining a set of demonstrated primitive skills.

\textbf{Local Motion Planner:} The authors’ recent work \cite{yc22} has developed the local motion planner, which is briefly introduced here without delving into technical details for the paper’s self-containment.

Let $\boldsymbol{DP} = \{\boldsymbol{D}_1, \boldsymbol{D}_2, \ldots, \boldsymbol{D}_n\}$ represent a demonstrated primitive skill in the task-space, where $\boldsymbol{D}_i$ is a dual quaternion representation for the $i^{\text{th}}$ configuration in the time sequence during the motion. The transformation, $\boldsymbol{\delta}_i$, between the last pose and every other pose is:

\begin{equation}
    \boldsymbol{\delta}_i = \boldsymbol{D}_{(i-1)}^* \otimes \boldsymbol{D}_n, \quad i = 2, \ldots, n
\end{equation}

where $\otimes$ represents dual quaternion multiplication and $\boldsymbol{D}^*$ denotes the conjugate of $\boldsymbol{D}$. Thus, the sequence of $\delta_i$ represents a sequence of transformations. Note that all implicit task constraints in a human demonstration are embedded in the sequence of $\boldsymbol{\delta}_i$ during the motion. This sequence can represent the features or semantics of human demonstrations. The feature of the $k^{\text{th}}$ human demonstration in the task space can be represented in a time sequence as:

\begin{equation}
\boldsymbol{HD}_k = \{\boldsymbol{\delta}_2^{(\boldsymbol{HD}_k)}, \ldots, \boldsymbol{\delta}_n^{(\boldsymbol{HD}_k)}\}
\end{equation}

\begin{figure}
    \centering
    \includegraphics[width=1\linewidth]{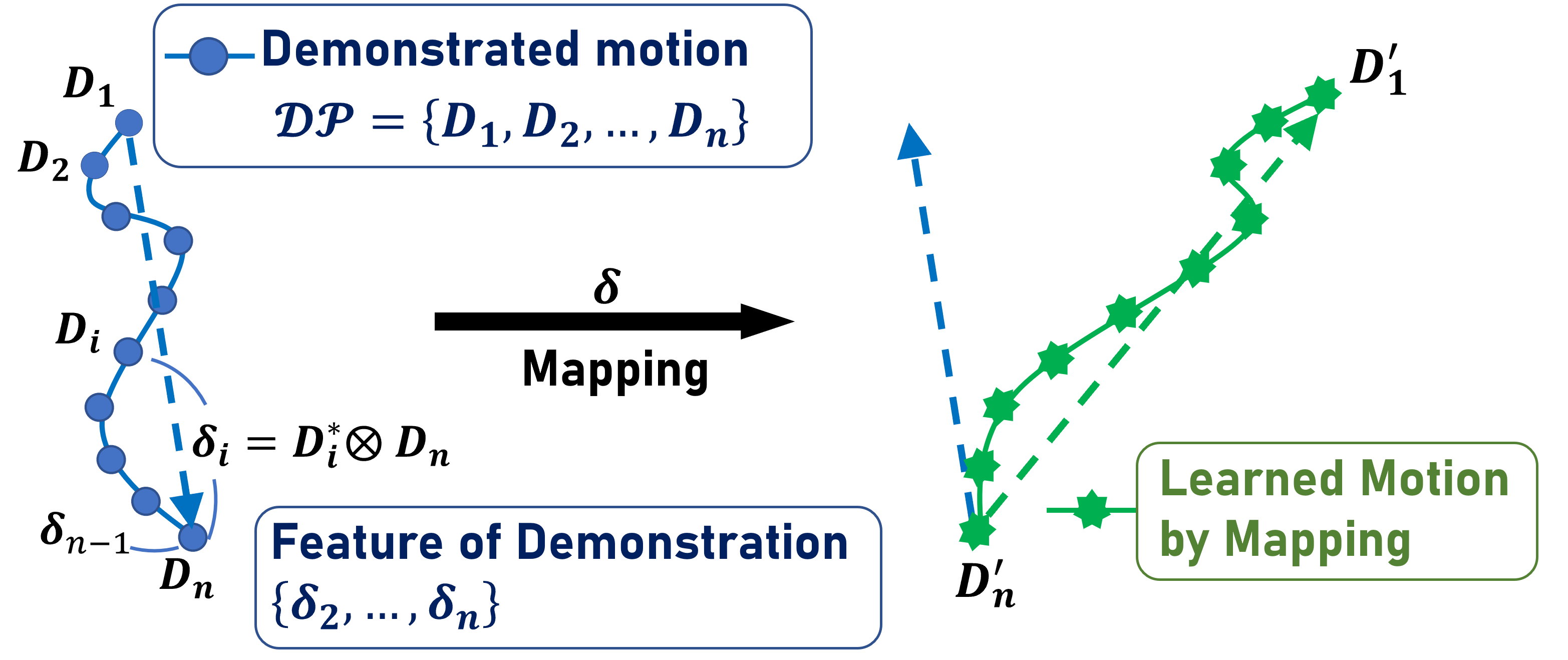}
    \caption{Mapping the demonstrated skill (blue line) to the new task with the new starting and goal configuration $D_1'$ and $D_n'$.}
    \label{fig:demo_mapping}
\end{figure}

For a new task, $t_k$, a mapping operation, $mp_{\boldsymbol{HD} \rightarrow t_k}$, is developed, which can align and enforce the feature of the demonstration to the task by using the quaternion sandwich operation \cite{yc22} ans is shown in Fig. \ref{fig:demo_mapping}.

\textbf{Global Planner:} However, manufacturing tasks such as assembly tasks are often complicated, which may include combinations of various primitive skills. Therefore, a library of $\boldsymbol{h}$ demonstrated primitive skills can be formed as:

\begin{equation}
  \boldsymbol{LB} = \{\boldsymbol{HD}_1, \boldsymbol{HD}_2, \ldots, \boldsymbol{HD}_h\}   
\end{equation}

For any task instance, a robot should be able to look at the library of demonstrations and be able to learn and combine the most suitable demonstrations by using the local motion planner. To do this, one may need to go through all possible subsets of new tasks and evaluate each $\boldsymbol{HD}_k$, $k = 1, 2, \ldots, h$, in the library $\boldsymbol{LB}$, which is an NP-hard problem. The state space of the problem would be huge if the constraints of new tasks and the number of demonstrated skills are large. The problem can be formulated as a model-free reinforcement learning problem in the Markov Decision Process (MDP) framework. In the context of the robot learning global planning, the RL problem is a tuple $\langle S, A, R, S' \rangle$, and is formulated as follows.

The state space $S$ contains a set of system states, which include the current configuration of the end-effector and all possible task segments. Let $s_t$ denote the system state at time $t$, $s_t \in S$, then $s_t$ is defined as

\begin{equation}
    s_t = \langle \text{EE}_t, \text{ta}_t \rangle
\end{equation}

where $\text{EE}_t$ is the current configuration of the robot’s end-effector at $t$, $\text{ta}_t \subseteq TK$ is the possible task segments at $t$.
The action space $A$ is the set of action pairs for the robot learner to decide on how to segment a new task and what demonstrations should be selected. The action $a_t \in A$, can be defined as

\begin{equation}
a_t = \langle \text{ts}_t, \boldsymbol{HD}_{(t,l)} \rangle 
\end{equation}

where $\text{ts}_t \in \text{ta}_t$ and $\boldsymbol{HD}_{(t,l)} \in \boldsymbol{LB}$.

The reward function is determined based on the Euclidean distance between the feature of the demonstration, $ \boldsymbol{\delta}_i^{(\boldsymbol{HD}_i)}$, and the feature of a new task, $ \boldsymbol{\delta}_l^{(\text{ts}_t)}$, as in \cite{yc22}. Let $\Delta_\beta$ be a predefined tolerance value, then:
\begin{equation}
r_t = \begin{cases}
-\sum_{l=j}^n \beta( \boldsymbol{\delta}_i^{(\boldsymbol{HD}_i)}, \boldsymbol{\delta}_l^{(\text{ts}_t)}) &~ \text{if } ~\beta( \boldsymbol{\delta}_i^{(\boldsymbol{HD}_i)},  \boldsymbol{\delta}_l^{(\text{ts}_t)}) \leq \Delta_\beta \\
-\infty &~ \text{otherwise}
\end{cases}
\end{equation}

\begin{figure}
    \centering
    \includegraphics[width=0.75\linewidth]{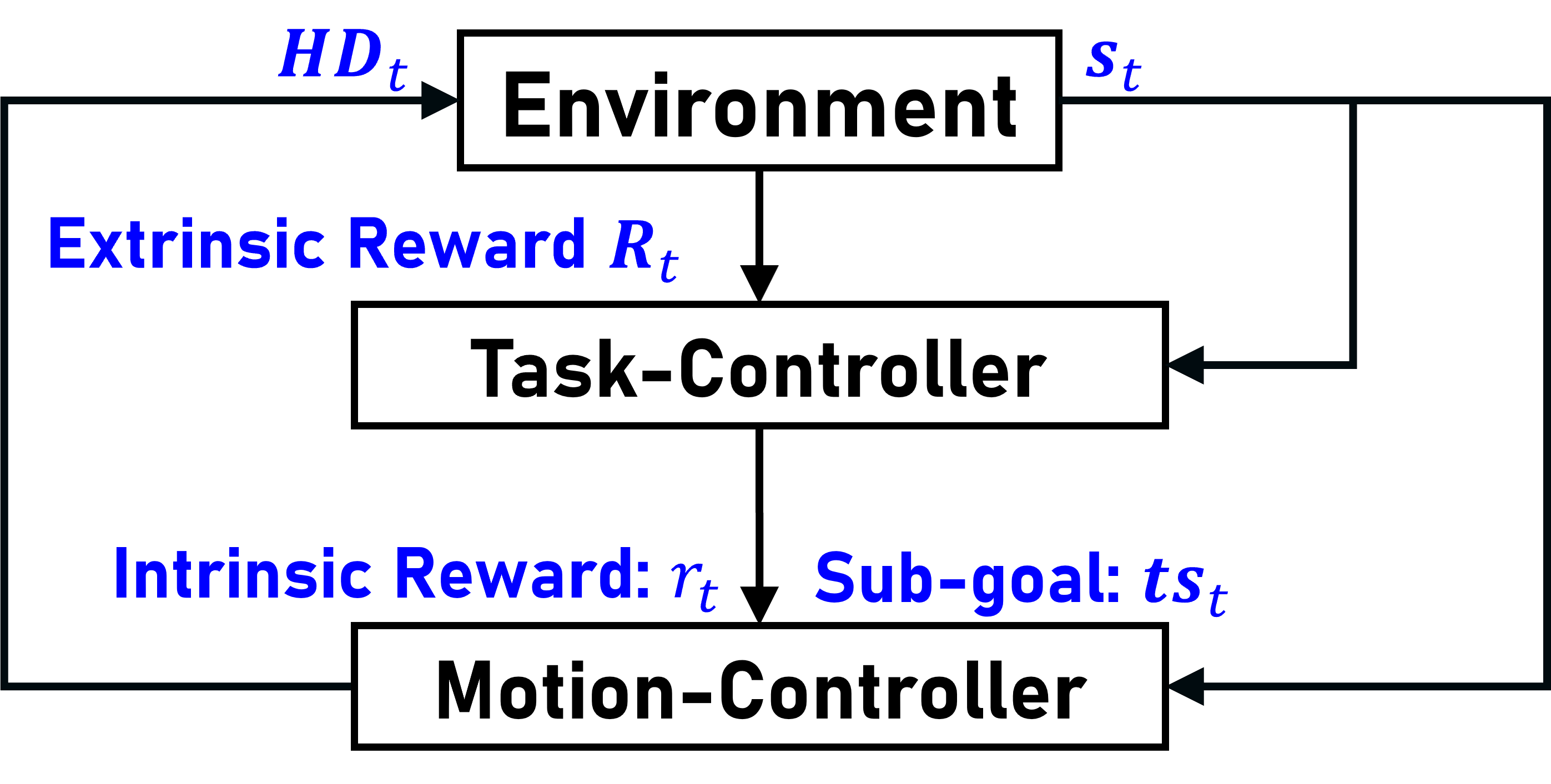}
    \caption{Human-in-the-loop LfD based RL scheme}
    \label{fig:HRL_Lfd}
\end{figure}

The standard RL approach needs to simultaneously determine the action pair $\langle \text{ts}_t, \boldsymbol{HD}_{(t,l)} \rangle$, and the state-action space grows exponentially with the number of features and task-relevant constraints. To alleviate the curse of dimensionality, an RL scheme is developed (see Fig. \ref{fig:HRL_Lfd}), where the agent uses a two-level hierarchy consisting of a task-controller and a motion-controller with two inter-dependent networks.

The task-controller is to find a policy that specifies the subgoal, $\text{ts}_t$, under the global state $s_t$. The task-controller will determine how to segment the new task through estimating the value function $Q(s_t, \text{ts}_t)$, such that the extrinsic reward, $R_t$, can be maximized. The $Q(s_t, \text{ts}_t)$ function of the task-controller is estimated as:
\begin{equation}
    Q(s_t, \text{ts}_t) = \mathbb{E}_{\pi_Q} [R_t \mid s_t = s, \text{ts}_t = ts]
\end{equation}

where $\pi_Q$ is the global policy over subgoals, $R_t$ is the extrinsic reward for the meta-controller and is defined as:
\begin{equation}
    R_t = \sum_{t'=0}^t r_t
\end{equation}

where $r_t$ is the intrinsic reward for the next level critic-controller.

The objective of the task-controller is to find an optimal policy, $\pi_{\text{ts}}^*$, such that $R_t$ can be maximized. $\pi_{\text{ts}}^*$ can be defined as:

\begin{equation}
    \pi_Q^*(\text{ts} \mid s) = 
\begin{cases} 
1, & \text{if } \text{ts} = \underset{\text{ts}_t \subseteq TK}{\max} \{Q(s_t, \text{ts}_t)\} \\
0, & \text{otherwise}
\end{cases}
\end{equation}
The motion-controller is to continue to determine the policy on specifying the action $\boldsymbol{HD}_{(t,l)}$ under the global state $s_t$ and the current subgoal, $\text{ts}_t$, that is flown down from the task-controller. The motion-controller will decide pertinent demonstrated primitive skills by estimating a value function, $q(s_t, \text{ts}_t, \boldsymbol{HD}_{(t,l)})$, so that the intrinsic reward, $r_t$, can be maximized. The value function can be written as:
\begin{equation}
    q(s_t, \text{ts}_t, \boldsymbol{HD}_{(t,l)}) = \mathbb{E}_{\pi} [r_t \mid s_t = s, \text{ts}_t = ts, \boldsymbol{HD}_{(t,l)} = \boldsymbol{HD}_l]
\end{equation}

where $\text{ts}$ is the given subgoal from the task-controller in state $s$, and $\pi_q$ is the policy on how to select human demonstrations.
The intrinsic reward, $r_t$, is to compare semantic similarity between the human demonstrations, $\boldsymbol{HD}_{(t,l)}$, and the subgoal, $\text{ts}_t$, using Eq. (11). The internal critic checks if the subgoal is reached and provides an appropriate intrinsic reward to the controller. The optimal policy $\pi_q^*$ of the critic-controller is defined as:

\begin{equation}
    \pi_q^*(\boldsymbol{HD}_l \mid s, \text{ts}) = 
\begin{cases} 
1, & \text{if } \boldsymbol{HD}_l = \underset{\boldsymbol{HD}_{(t,l)} \in \boldsymbol{LB}}{\arg \max} \{q(s_t, \text{ts}_t, \boldsymbol{HD}_{(t,l)})\} \\
0, & \text{otherwise}
\end{cases}
\end{equation}

The algorithm of offline training and online execution of the RL-LfD method is shown in Algorithm  \ref{alg:hrl_lfd}.

\begin{algorithm}
\caption{RL-LfD Method: Offline Training and Online Execution}
\label{alg:hrl_lfd}
\begin{algorithmic}[1]
\Procedure{Offline Training of the RL-LfD Method}{}
\State \textbf{Input:} TK, \(\boldsymbol{LB}\)
\State Initialize \( H(s, ts) \) and \( q(s, ts, HD_l) \) randomly
\State Initialize \( s_0 \) with \( EE_0 \) and \( ta_0 \)
\For{$t = 1, \ldots, T$}
    \For{$t' = 1, \ldots, \tau$}
        \State Compute \( r_t \) using Eq. (10)
        \State Update \( q_{t'}(s_t, ts_t, HD_{t', l}) \) using Eq. (14)
    \EndFor
    \State Update \( \tilde{r}_t \leftarrow \sum_{t' = 0}^t r_{t'} \) and update \( H_t(s_t, ts_t) \)
\EndFor
\State \textbf{Output:} \( H(s, ta) \) and \( q(s, ta, HD_l) \)
\EndProcedure

\Procedure{Online Execution of the RL-LfD Method}{}
\State \textbf{Input:} TK, \(\boldsymbol{LB}\), \( H(s, ta) \), and \( q(s, ta, HD_l) \)
\While{$ EE_t $ is not the last configuration in TK}
    \State Select the task segment \( ts = \arg\max_{ts_t \subseteq TK} Q(s_t, ts_t) \)
    \State Select the demonstration \( HD_l = \arg\max_{HD_{t, l} \in \boldsymbol{LB}} q(s_t, ts_t, HD_{t, l}) \)
    \State Mapping \( mp_{HD_l \rightarrow ts} \) to calculate \( traj_{LfD} \)
    \State Update \( EE_t \)
\EndWhile
\State \textbf{Output:} \( traj_{LfD} \)
\EndProcedure
\end{algorithmic}
\end{algorithm}
\section{Joint-Space Feasibility DRL Motion Planner}\label{sec:drl_pla}
As mentioned earlier, in cases where the RL-LfD trajectory enters an infeasible region, three possible scenarios may unfold: the manipulator's end effector may fail to reach a position, encounter self-conflict or workspace collisions, or experience reduced manipulability below a set tolerance, which constitutes a joint-space failure. To address this issue, the authors build upon our recent work \cite{ko22} by applying a Deep Reinforcement Learning (DRL) approach. The motion planning problem is formulated as an MDP, and Proximal Policy Optimization algorithm is used in the offline training of the joint space motion planner. PPO belongs to the family of model-free, on-policy Deep RL algorithms, which achieves strong performance across a wide range of continuous and discrete control tasks while maintaining stability and ease of implementation. After training, this motion planner is then implemented during real-time execution to rectify any unfeasible segments of the LfD trajectory. 

\subsection{MDP Formulation}
Given the workspace, the robot and the target position, the primary components of the MDP can be defined as follows. The state space \( S \) includes all information regarding the robot and the environment conditions. The state \( s_t \in S \) of the RL agent is defined as:

\[
s_t = \langle JP_t, JO_t, LV_t, AV_t, TP_t, TO_t, RL_t \rangle \tag{16}
\]

where \( JP_t, JO_t \in \mathbb{R}^{3 \times n} \), \( n \) is the degree of freedom, denotes the x, y, z positions and Euler angles of each joint, respectively; \( LV_t, AV_t \in \mathbb{R}^{3 \times n} \) denotes the linear and angular velocities of each joint, respectively; \( TP_t, TO_t \in \mathbb{R}^3 \) denotes the x, y, z positions and Euler angles of the end-effector, respectively; \( RL_t \in \mathbb{R}^{25} \) denotes the length of the ray that generated from the end-effector to the surface of the environment. This ray trace can reflect the real-time environment conditions. In this joint space motion planner, 25 rays are generated in different angles to the end-effector using a PyBullet library.

The action of the RL agent is defined as a vector of joint angles:

\begin{equation}
a_t = [\theta_{1t}, \theta_{2t}, \ldots, \theta_{nt}] 
\end{equation}

where \( \theta_{it}, i \in 1, \ldots, n, n \) is the DOF, denotes the joint angle, which is limited within \([-1, 1]\). At every time instance \( t \), \( a_t \) depicts a robot's configuration. The feasibility of this configuration plays a pivotal role in determining the reward. Moreover, navigating the robot towards the target region requires considering the gap between the end-effector and the intended goal as a significant reward component. The position of the end-effector can be calculated using forward kinematics \( FK: a_t \rightarrow D_t \). Let \( d_t \) represent the distance between the end effector and the object, and let \( r \) signify the radius of the target region relative to the goal position. Consequently, the reward for the PPO agent is described as:

\begin{equation}
\label{eq:total_reward}
R_t = \begin{cases} 
fea(D_t) - d_t, & \text{if } d_t \geq r \\
0.1, & \text{otherwise}
\end{cases} 
\end{equation}

\begin{algorithm}
\caption{DRL Method: Offline Training and Online Execution}
\label{alg:drl}
\begin{algorithmic}[1]
\Procedure{DRL Method}{}
\State \textbf{Input:} $\boldsymbol{WS}$, robot, starting and goal positions, radius of the target area \( r \), initial neural network weights \( \theta \)
\While{the episode does not terminate}
    \State Observe the state \( s \) and select the action \( a \sim \pi_{\theta}(\cdot|s) \)
    \State Execute \( a \) in the environment and get the next state \( s' \)
    \If{ \( s' \) is a feasible state}
        \State Calculate \( R_t \) using Eq. \eqref{eq:total_reward}
        \State Store \( (s, a, r, s', \text{done}) \) in replay buffer \( D \)
    \EndIf
    \If{it is time to update the target neural network}
        \State Randomly sample a batch of transitions from \( D \)
        \State Compute the loss function and update the policy
        \State Update the target network’s \( \theta \)
    \EndIf
\EndWhile
\State \textbf{Output:} target network \( \theta \)
\EndProcedure
\\
\Procedure{Online Execution of the DRL Method}{}
\State \textbf{Input:} Robot starting pose, Goal Position, Neural Network \( \theta \)
\While{end-effector position is not in the target area}
    \State Execute the action \( a_t \sim \pi_{\theta^*}(\cdot|s_t) \)
    \State Append \( a_t \) to \( \text{traj}_{DRL} \)
    \State Update the current end-effector pose
\EndWhile
\State \textbf{Output:} \( \text{traj}_{DRL} \)
\EndProcedure
\end{algorithmic}
\end{algorithm}
\subsection{Online Execution of the Joint Space Motion Planner}

It is noted that by leveraging the RL-LfD, the DRL method does not need to learn from scratch by searching the whole workspace. It only needs to learn the infeasible segments. After training, the trained policy, \( \pi^* \) (weights of neural networks) with the optimal average reward is used for the online execution.

For online execution, given the infeasible segment, \( \neg FJ_{ij} = \{D_i, D_{i+1}, \ldots, D_j\} \), of the RL-LfD trajectory, the starting and goal pose of the joint space motion planner is \( D_{i-1} \) and \( D_{j+1} \). Using the pair \(\{D_{i-1}, D_{j+1}\}\) as the input, the output of the joint space motion planner is a feasible trajectory \( \text{traj}_{DRL} \) that moves the robot from \( D_{i-1} \) to \( D_{j+1}' \). The algorithm of offline training and online execution of the DRL method is shown in Algorithm  \ref{alg:drl}.

\section{Experiments}\label{sec:exp}
To validate our proposed method and to gain an in-depth understanding about the efficacy, we carry out multiple experiments, both in simulation and on a physical setup. We also discuss a case study in which the proposed method was used for conducting Non-Destructive Testing in a real-world industrial setup. The performance of the proposed hybrid robot motion planning method is evaluated using two metrics: (1) computing time to achieve a steady motion planning policy in the offline training; (2) the success rate of the motion plan in the online execution. To assess the performance of the proposed method, one purely DRL-based robot motion planning method \cite{ko22} is used for comparison. Based on the results of the case study, three key conclusions can be drawn: (1) the proposed hybrid robot motion planning method is effective in generating adaptive trajectories for different tasks; (2) in offline training, the proposed hybrid robot motion planning method outperforms the purely DRL-based method in training efficiency and generating feasible configurations; (3) when provided with same training time, the proposed hybrid robot motion planning method outperforms the purely DRL-based method in the success rate of achieving different tasks.

\subsection{Simulation Study}
We first validate our proposed method in a simulation environment. For the purpose of the experiments we utilize a Kinova Gen-3 7 DoF manipulator in the PyBullet simulation environment. All the models are trained using a Nvidia RTX 3090 GPU with a AMD Ryzen Threadripper processor with 32 Cores and 256 GB RAM and 2 TB SSD. 

\begin{figure}
\begin{subfigure}{\textwidth}
  \centering
  \includegraphics[width=0.9\linewidth]{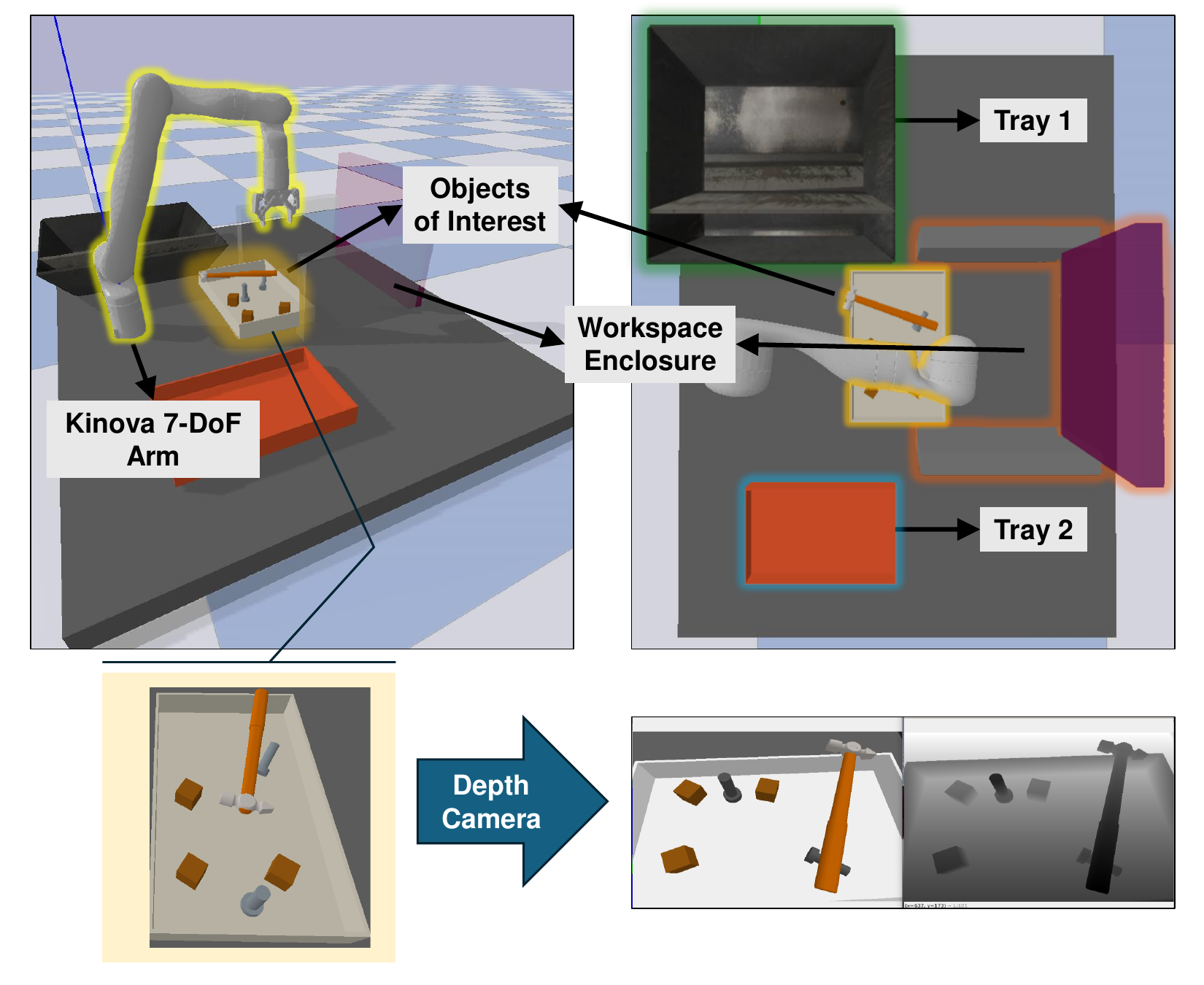}
  \caption{Experimental setup for simulation study task used in PyBullet.}
  \label{fig:exp_setup_py}
\end{subfigure}
\begin{subfigure}{\textwidth}
  \centering
  \includegraphics[width=0.5\linewidth]{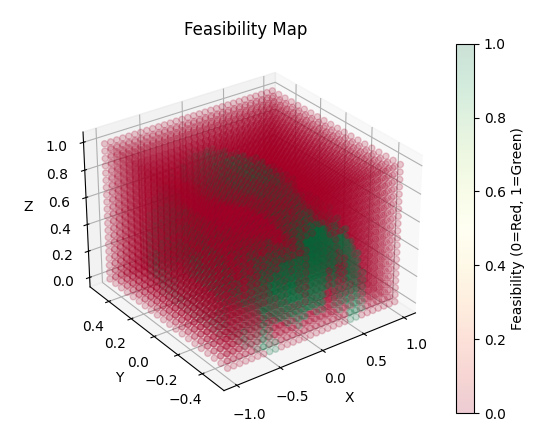}
  \caption{Feasibility map of the workspace used in the simulation study.}
  \label{fig:Feas_map_sim}
\end{subfigure}
\caption{Experimental setup with feasibility map used for simulation study.}
\label{fig:exp_setup}
\end{figure}

The robot is tasked with completing a sequence of tasks simulating material handling in an enclosed workspace with obstacles as shown in Fig. \ref{fig:exp_setup_py}. We use a Kinova Gen-3 7-DoF manipulator with Robotiq 2f-85 end-effector to complete all the tasks. Additional to the trays and objects in the scene, the robot also has an enclosure which can further add to more collisions. Figure \ref{fig:Feas_map_sim} shows the feasibility map ($\boldsymbol{FR}$) of the workspace.



\subsubsection{Training}
The three agents, namely the RL-LfD agent, the DRL-Feasibility agent and the RL Switching agent are all trained separately. First the RL-LfD agent and the DRL-Feasibility agents are trained separately offline. Eventually, the RL-Switching agent is trained online to learn an efficient way to combine the two agents given the scenario. We utilize a Deep Q-Network (DQN) to train the RL-LfD agent. The state is defined as the sequence of poses and the actions are defined as the pair of sub-task segment and skillset as discussed in Sec. \ref{sec:hrl_lfd}. Table \ref{tab:DQN} shows the model parameters of the DQN.

\begin{table}[htbp]
\caption{DQN Hyperparameters (RL-LfD Agent)}
\centering
\label{tab:DQN}
\begin{tabular}{l|c}

\hline
\textbf{Parameter}         & \textbf{Value} \\ \hline \hline
Batch Size (BATCH\_SIZE)   & 256            \\ 
Discount Factor ($\gamma$) & 0.92           \\ 
Start Exploration Rate ($\epsilon_{start}$) & 0.95 \\ 
End Exploration Rate ($\epsilon_{end}$) & 0.05     \\ 
Exploration Decay (EPS\_DECAY) & 2000      \\ 
Soft Update Factor ($\tau$) & 0.005       \\ 
Learning Rate ($\alpha$)          & $1 \times 10^{-4}$ \\
\hline
\end{tabular}
\end{table}

\begin{figure}
    \centering
    \includegraphics[width=1\linewidth]{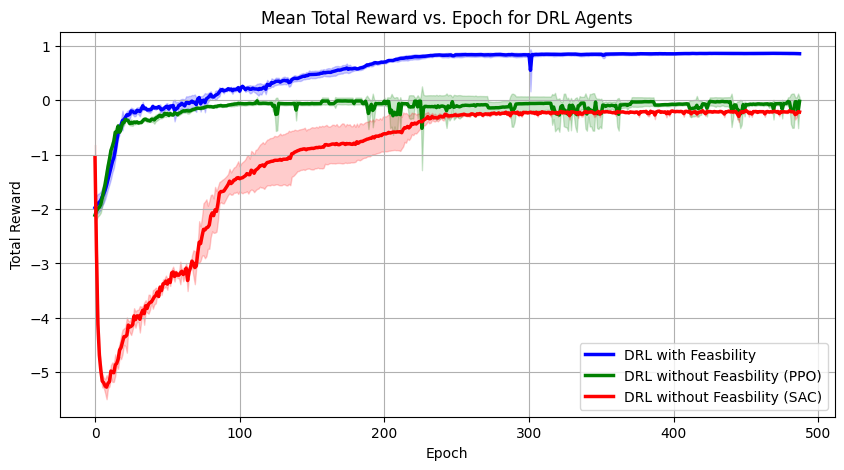 }
    \caption{Plot shows comparison between Total Reward v/s Steps for the Proposed Feasibility DRL and Simple DRL method.}
    \label{fig:training_comp}
\end{figure}

Since we include the feasibillity in the reward function of the DRL agent we study the comparative performance in training against DRL agents which do not include the feasibility feedback \cite{ko22}. We refer to our agent as the DRL-Feasibility agent henceforth. The DRL-Feasibility agent is trained using the Proximal Policy Optimization (PPO). While PPO is an online learning method, we utilize a batch-training approach to emulate off-policy training characteristics and further reducing the training time. We compare the training performance of the DRL-Feasibility agent with two pure DRL techniques, a PPO based agent and a SAC based agent. For both cases is the reward is defined as the distance from the goal pose in joint space similar to \cite{ko22}. Since the DRL-Feasibility extra reward for feasibility we compensate for it by providing an additional reward when goal is reached to maintain a similar scale of reward functions for all three cases.  
Fig. \ref{fig:training_comp} shows the training curve of the three cases. Each agent is trained for $10^6$ steps in batches of 2048 episode. The model parameters are updated in mini-batches of 64. Each epoch represents training after each batch update i.e. 2048 episodes. As it can be seen the DRL-Feasibility agent has a higher reward hence verifying our hypothesis. While the Pure DRL PPO agent does start to converge faster, the total reward plateaus rarely does it ever finish the task. For the Pure DRL SAC agent, the initial rise time in learning is the slowest, and the agent fails to reach the goal in training as the reward is always non-positive. Table \ref{tab:feas} shows the parameters of the three models respectively.

\begin{figure}
    \centering
    \includegraphics[width=1\linewidth]{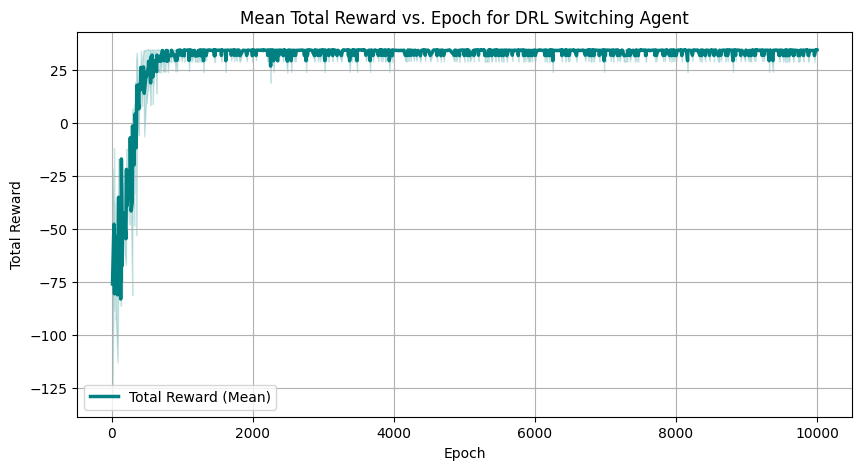 }
    \caption{Training curve for the RL Switching Agent.}
    \label{fig:switch_train}
\end{figure}

\begin{table}
    \centering
    \caption{PPO Model Parameters (DRL-Feasibility Agent)}
    \begin{tabular}{l|c}
    \hline
    \textbf{Parameter} & \textbf{Value} \\ \hline \hline
       Learning Rate  & 0.0003\\
       Discount $\gamma$  & 0.99\\
       Minibatch Size & 64\\
       Num Steps  & 2048\\
       Entropy Coefficient $c_1$  & 0.0\\
       VF Coefficient $c_2$  & 0.5\\
       Max Grad Norm & 0.5 \\
       \hline
    \end{tabular}
    \label{tab:feas}
\end{table}
Figure \ref{fig:switch_train} shows the training curve for the RL Switching agent. The Switching agent is trained using a discrete PPO algorithm. Discretized PPO is chosen to train this due due to it's ability to handle discrete state-action space in addition to the stability and speed in learning. Table \ref{tab:switch_ppo_params} shows the model parameters used. The model was trained for 10000 epochs in a batch update format which each epoch being 200 episodes long. The model parameters are updated in a mini-batch size of 32 steps.

\begin{table}
    \centering
    \caption{Discrete PPO Model Parameters (RL-Switching Agent)}
    \begin{tabular}{l|c}
    \hline
    \textbf{Parameter} & \textbf{Value} \\ \hline \hline
       Learning Rate  & 0.001\\
       Discount $\gamma$  & 0.96\\
       Minibatch Size & 32\\
       Num Steps  & 200\\
       Entropy Coefficient $c_1$  & 0.0\\
       VF Coefficient $c_2$  & 0.5\\
       Max Grad Norm & 0.5 \\
       \hline
    \end{tabular}
    \label{tab:switch_ppo_params}
\end{table}

\subsubsection{Execution}
We validate the trained model in execution for the task of placing objects from a enclosed workspace to it's respective trays. Figure \ref{fig:exp_setup_py} shows the setup of the scenario developed in PyBullet \cite{cb16}. In this scenario multiple objects are placed in front of the robot in an enclosed workspace. The robot must pick up each object and place it in the desired tray. The tools (hammer and bolts) go to the Tray 1 and other objects (blocks) go to Tray 2. The trays are of distinct size and are placed such that the robot will have to avoid collision multiple times. As discussed in Sec. \ref{sec:hrl_lfd} we define the task as a sequence of poses in SE(3). In this scenario, each sub-task is defined as the initial pose of the object and the final pose of the object. 

It must be noticed that picking up each object requires a particular set of skill, due to characteristics such as surface friction, center of mass, etc. As discussed in Sec. \ref{sec:hrl_lfd}, we leverage demonstrations for such dexterous sub-tasks. However, the DRL planner does not have access to these skill sets. 

\begin{figure}
    \centering
    \includegraphics[width=1\linewidth]{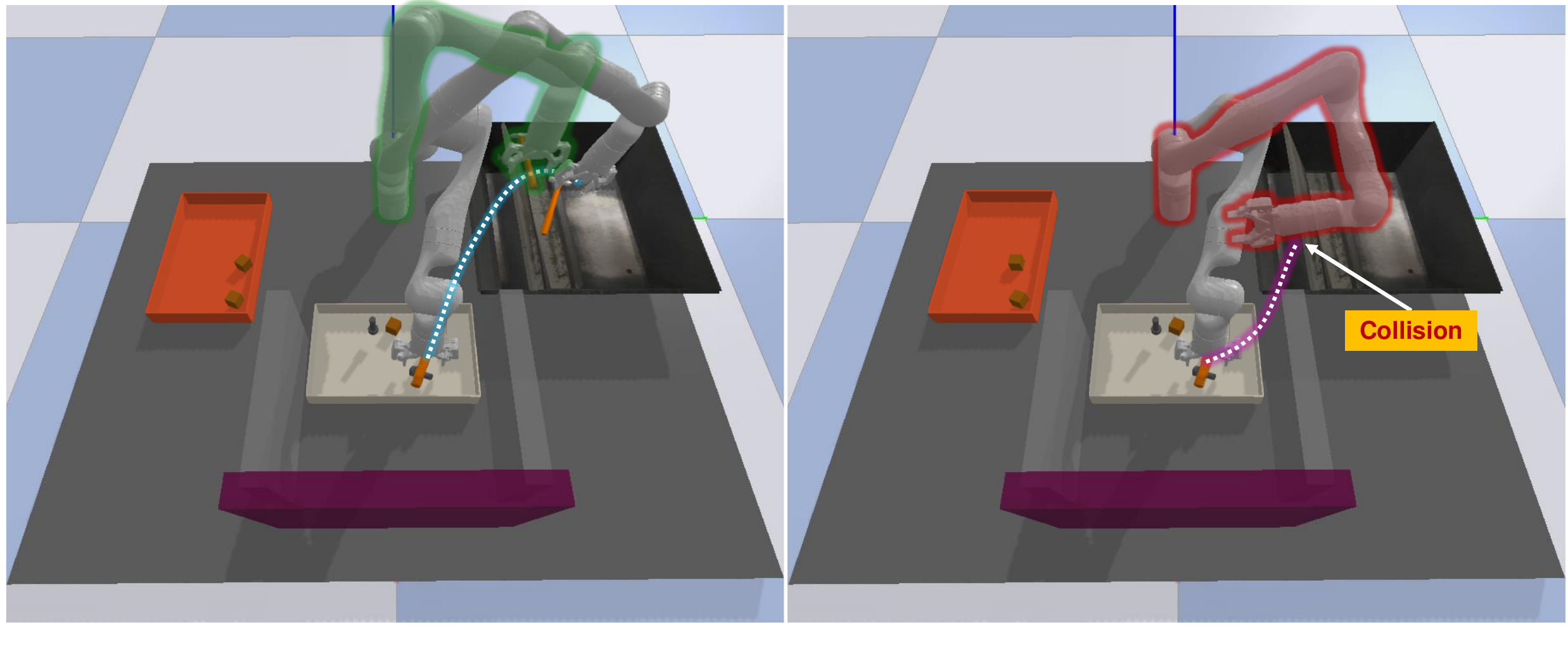 }
    \caption{Execution of pick and place task. Comparison between the Hybrid Learning approach (left) and the RL-LfD approach (right).}
    \label{fig:experiements}
\end{figure}

To validate our method's efficacy we compared the proposed Hybrid Learning method against DRL Planner (PPO) and just the RL-LfD method. We repeat the experiments 100 times and evaluated the success rate if all the items were placed at the desired location without causing any collisions. The proposed hybrid learning method has a success rate of 82\%, whereas the RL-LfD planner had a success rate of 0\% success rate and the DRL planner has 23\% success rate. The proposed method almost always succeeds to execute the motion plan without any collisions and while maintaining feasibility, however, during the switching between $\text{traj}_{LfD}$ and $\text{traj}_{DRL}$ caused the objects to drop out from the gripper or cause collisions. While the LfD has a very poor success rate when considering the complete task, it has slightly higher success rate when considering partial task, i.e. for certain sub-tasks (placing the blocks in Tray 2) generated collision free feasible trajectories but while placing the tools in Tray 1, following $\text{traj}_{LfD}$ always leads to collision. Lastly, the DRL planner was able to successfully complete the task on certain occasions it was mostly limited by the inability to maintain any gripping actions and would often lead to collisions hence highlighting the need for human demonstrations.

\subsection{Industrial Case Study}
This section presents simulated experiments conducted on a fluorescent penetrant inspection (FPI) task, which is the most widely used Non-Destructive Testing (NDT) method in the aerospace industry \cite{ko21}. The experimental setup for the FPI task is depicted in Fig. 5(a). The task to be performed is shown in Fig. 5(c), where the Fanuc LR Mate 200iD robot is required to move from the initial home configuration, \( \text{con}_1 \), to the hovering position, \( \text{con}_2 \), above the center of the tray, and then brush one blade from \( \text{con}_3 \) (an initial configuration on a blade) to \( \text{con}_4 \) (an ending configuration on a blade). A set of 20 fundamental skills commonly employed in FPI tasks are provided. Fig. 5(b) illustrates three examples of the demonstrated skills, including rotating, twisting, and translating on a Kinova Gen3 robot.

\begin{figure}
\begin{subfigure}{\textwidth}
  \centering
  \includegraphics[width=\linewidth]{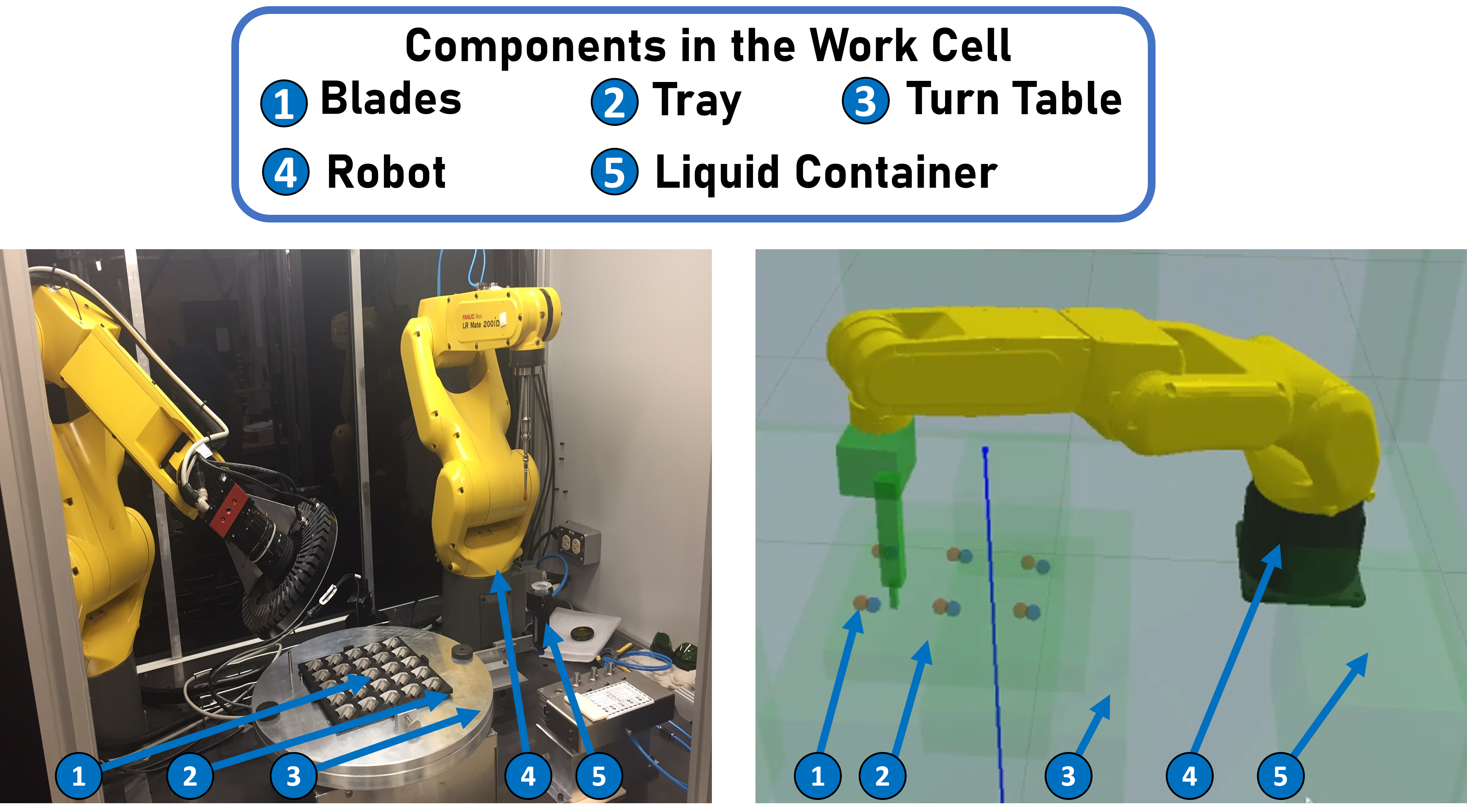}
  \caption{}
  \label{fig:sfig1}
\end{subfigure}
\begin{subfigure}{\textwidth}
  \centering
  \includegraphics[width=\linewidth]{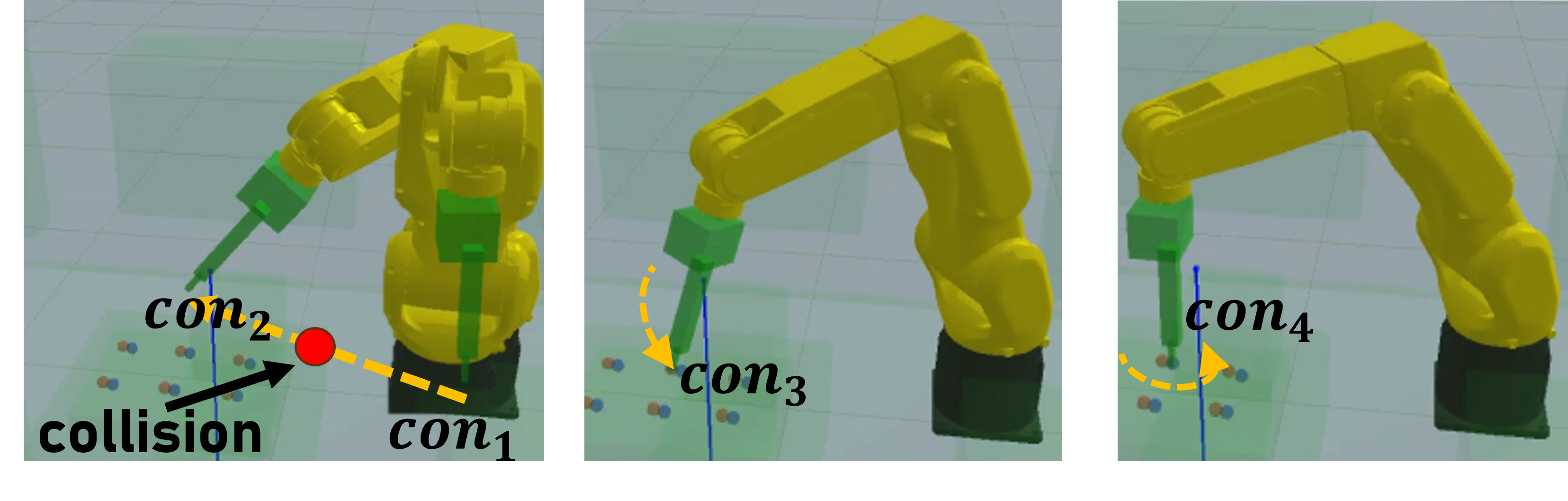}
  \caption{}
  \label{fig:sfig2}
\end{subfigure}
\begin{subfigure}{\textwidth}
  \centering
  \includegraphics[width=\linewidth]{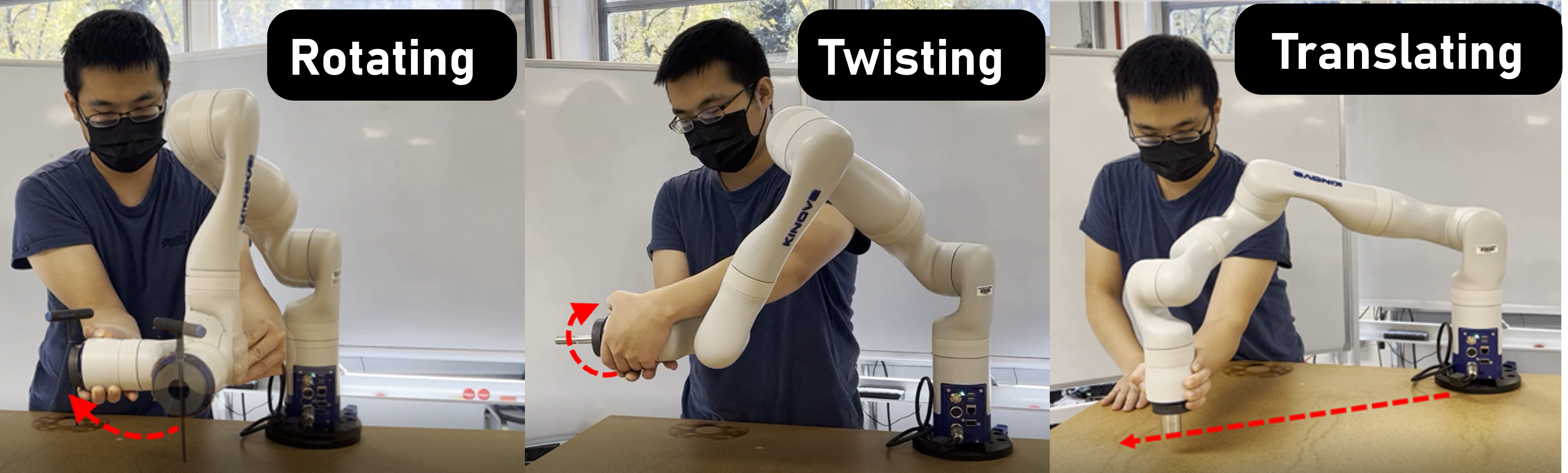}
  \caption{}
  \label{fig:sfig2}
\end{subfigure}
\caption{Fanuc LR Mate 200iD test bed and kinesthetic demonstrations. (a) Components in the FPI work cell. The simulated work cell (right) is set up in Pybullet, which is the digital twin of the real robotic work cell (left). (b) FPI tasks to be performed. (c) Kinesthetic demonstrations on a Kinova Gen3 robot.}
\label{fig:fig}
\end{figure}


\subsubsection{Offline Training of the Hybrid Motion Planning Method}
All offline training is executed on an 8-core workstation processor paired with an Nvidia GPU. For effective offline training, 100 FPI tasks are generated with randomly chosen starting and goal positions in the work cell. This diverse task set allows us to evaluate the system's performance across various scenarios. First, Algorithm  \ref{alg:hrl_lfd} is employed to train the task space RL-LfD motion planner. After training, the feasibility analysis is conducted to identify infeasible segments. These segments' starting and ending configurations are then utilized as input data to train the joint space DRL-Feasibility motion planner using Algorithm  \ref{alg:drl}. For comparison, a purely DRL-based method is trained within the entire workspace by following the same task constraints. During the offline training phase, the radius of the target region is set to a tolerance of 25 centimeters. Under these settings, the DRL-Feasibility-based method can learn a steady policy within an acceptable time frame. 

\begin{figure}
    \centering
    \includegraphics[width=1\linewidth]{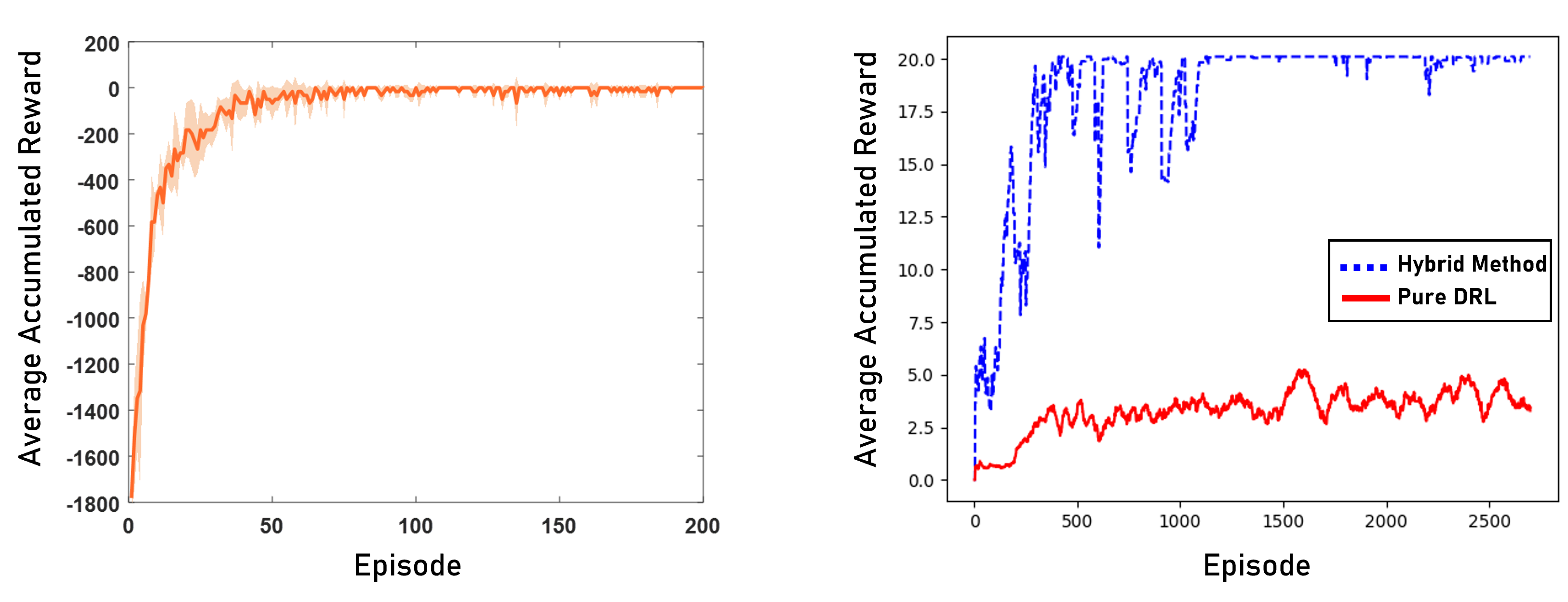}
    \caption{Offline training of the RL-LfD method (left) and the offline training of the DRL method (right).}
    \label{fig:Training_comp_ndt}
\end{figure}

As shown in the left plot of Fig. \ref{fig:Training_comp_ndt}, after around 100 episodes, the RL-LfD motion planner can obtain a steady policy that optimizes the average accumulated reward. This high training efficiency is achieved as the proposed RL-LfD method learns the kinematic features, which remain consistent across similar tasks despite variations in positions and orientations. In the right plot of Fig. \ref{fig:Training_comp_ndt}, the learning curve of DRL-Feasibility in the hybrid approach reaches a stable policy after around 1250 episodes. In contrast, the purely DRL method struggles to converge. Furthermore, the average accumulated reward obtained using the purely DRL approach is notably lower compared to the hybrid method. This discrepancy can be attributed to the inherent difficulty of the purely DRL method in searching for feasible configurations, ultimately leading to reduced rewards. The delayed progress in training of the purely DRL method stems from its learning from scratch, where it searches the feasible motion planning policy within the whole space to comprehend position and orientation task constraints. The hybrid approach, however, leverages the strengths of both task space RL-LfD and joint space DRL. The trained RL-LfD policy plans the trajectory based on the task-level understanding and the demonstrated features of skills, which does not need to be retrained when tasks change. In addition, the knowledge of the infeasible segments transferred from the RL-LfD method, and the feasibility map significantly reduces the searching space of the DRL-Feasibility based method and avoid learning from scratch, resulting in the high efficiency and adaptivity in the offline training. 

\subsubsection{Online Execution of the Hybrid Motion Planning Method }
\begin{figure}
    \centering
    \includegraphics[width=1\linewidth]{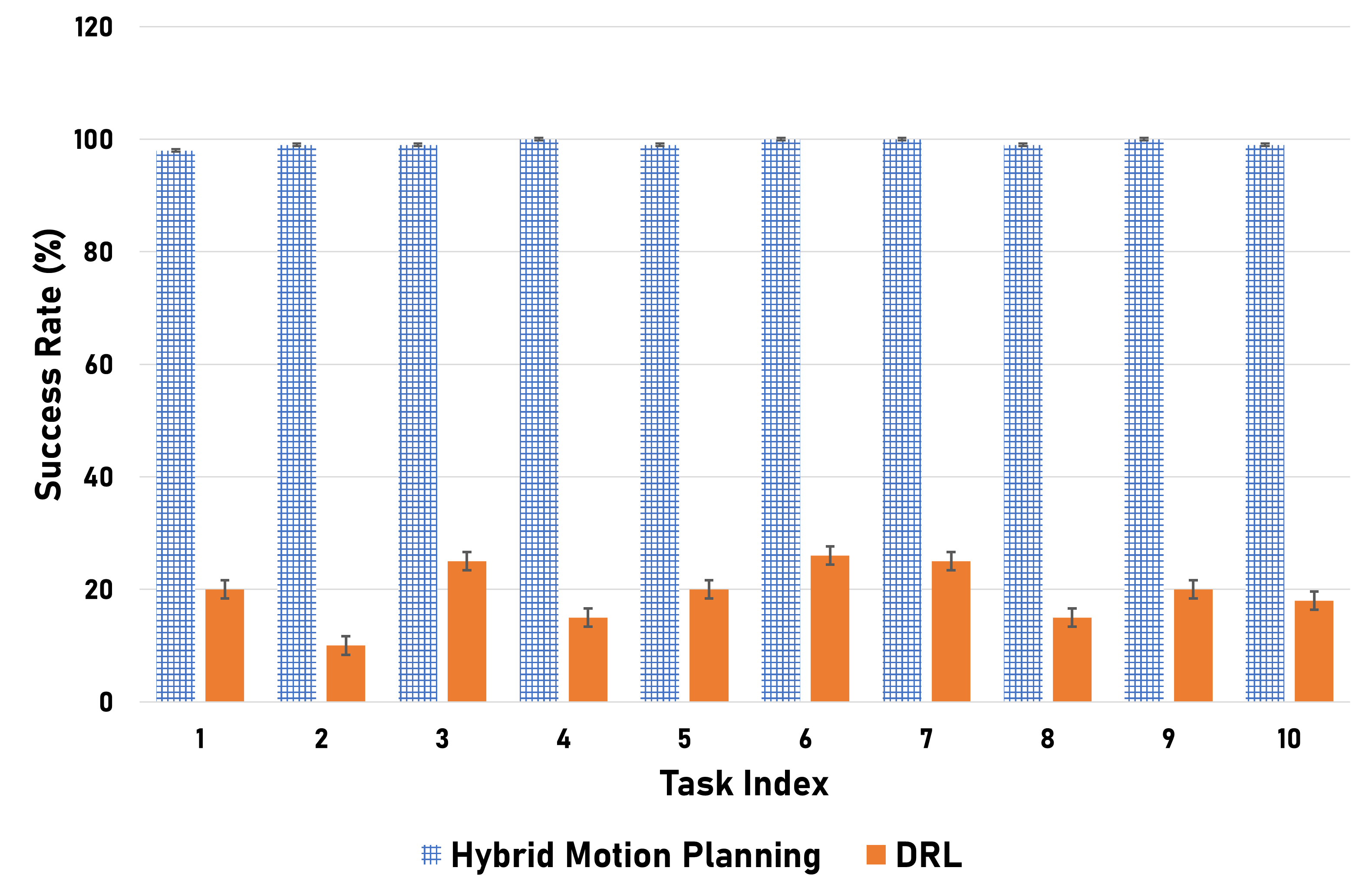}
    \caption{Online execution of the proposed hybrid motion planning method and the purely DRL method.}
    \label{fig:Online_comp}
\end{figure}

Trained policies from both the proposed hybrid method and the purely DRL-based method are employed to execute ten distinct painting tasks on ten blades. For each task, a total of 100 execution trials are conducted. A trial is deemed successful when the robot can reach all essential configurations, covering both position and orientation. This includes reaching the container, as well as the initial and ending configurations on each blade, all within a 5-centimeter tolerance for positions and a 5-degree tolerance for each Euler angle in the critical orientation. Fig. \ref{fig:Online_comp} illustrates the success rates of both approaches, highlighting the significant superiority of the hybrid method in successfully completing all tasks compared to the purely DRL approach. This contrast arises from the fact that DRL-Feasibility-based training relies on a less stringent tolerance for achieving a target configuration to establish a relatively stable policy. Conversely, the hybrid method capitalizes on RL-LfD to improve accuracy within the task space. 
In conclusion, the proposed hybrid motion planning method demonstrates both effectiveness and efficiency in generating feasible trajectories that satisfy task requirements, thereby providing significant benefits over methods solely based on DRL.

\section{Conclusion \& Future Work}\label{sec:conc}

This paper presents a multi-level hybrid robot motion planning method that integrates the strengths of task-space RL-LfD and joint-space DRL-based motion planning while minimizing their inherent limitations through an efficient switching mechanism. By training the DRL-based method using identified infeasible segments of the RL-LfD trajectory, the proposed approach significantly improves training efficiency and ensures the generation of a feasible motion planning policy. The RL-based switching mechanism further reduces the need for extensive heuristic tuning, enabling seamless execution.

The proposed method was validated in both simulation and real-world industrial environment. The validation results demonstrate that the hybrid learning framework leads to substantial improvements in both training time and execution accuracy. These findings suggest that, given a sufficient skill set and by training a DRL agent to emphasize feasibility, human demonstrations and deep learning techniques can be effectively integrating to enhance adaptability in autonomous robotic operations within condensed workspaces.

Future work will focus on improving the scalability and adaptability of the hybrid motion planning method by incorporating robot dynamics. Additionally, we aim to explore online learning strategies to enable multiple robots to adapt to dynamic work environments in real time.

\section{Acknowledgment}
\label{app1}
We would like to acknowledge the support of ARM Institute, General Motors and GE Research, for the work carried out in this manuscript. This work is supported by the National Science Foundation Grant CMMI 2243930.





\bibliographystyle{elsarticle-num} 
 \bibliography{ReferencesNew}
\end{document}